\documentclass[11pt]{article}

\usepackage[preprint]{acl}

\usepackage{times}
\usepackage{latexsym}

\usepackage[T1]{fontenc}

\usepackage[utf8]{inputenc}

\usepackage{microtype}

\usepackage{inconsolata}

\usepackage{graphicx}

\usepackage{hyperref}
\usepackage{url}
\usepackage{graphicx}
\usepackage{booktabs}      
\usepackage{amsmath,amssymb}
\usepackage{enumitem}      
\usepackage{float}         
\usepackage{pifont}        
\usepackage{xspace}
\usepackage{adjustbox}

\usepackage{soul}
\usepackage{xspace}
\usepackage[most]{tcolorbox}

\usepackage{xcolor}
\usepackage{colortbl}

\definecolor{sbPrimary}{HTML}{0F4C5C}   
\definecolor{sbPrimaryB}{HTML}{2A7387}  
\definecolor{sbPrimaryC}{HTML}{5C9FB0}  
\definecolor{sbPrimaryD}{HTML}{A5CBD5}  
\definecolor{sbAccent}{HTML}{E09F3E}    
\definecolor{sbAccentB}{HTML}{B87333}   

\definecolor{sbInk}{HTML}{14181A}
\definecolor{sbMuted}{HTML}{6B7A80}
\definecolor{sbRule}{HTML}{DFE5E7}
\definecolor{sbPale}{HTML}{EDF1F2}

\newcommand{\best}[1]{\textbf{\textcolor{sbPrimary}{#1}}}
\newcommand{\second}[1]{\underline{#1}}
\newcommand{\weak}[1]{\textcolor{sbAccentB}{#1}}
\newcommand{\headrow}{\rowcolor{sbPale}}

\newif\ifdraftmode
\draftmodetrue        

\ifdraftmode
  
  \newcommand{\todo}[1]{\textcolor{red}{\textbf{[TODO: #1]}}}

\else
  
  \newcommand{\todo}[1]{}
\fi

\hypersetup{
  colorlinks=true,
  linkcolor=sbPrimary,
}

\tcbuselibrary{listings,breakable,skins}

\tcbset{
  sbprompt/.style={
    enhanced,
    breakable,
    colback=sbPale!35!white,
    colframe=sbPrimary,
    colbacktitle=sbPrimary,
    coltitle=white,
    fonttitle=\small\bfseries,
    boxrule=0.4pt,
    arc=1.5mm,
    left=2mm,
    right=2mm,
    top=2mm,
    bottom=2mm,
    before skip=10pt,
    after skip=10pt,
    listing only,
    listing options={
      language={},
      basicstyle=\ttfamily\footnotesize,
      breaklines=true,
      breakatwhitespace=true,
      columns=fullflexible,
      keepspaces=true,
      showstringspaces=false,
      tabsize=2,
      breakindent=0pt,
      breakautoindent=false,
      escapeinside={(*@}{@*)},
    }
  }
}

\newtcblisting{promptbox}[1]{
  sbprompt,
  title={#1}
}

\newcommand{\bench}{\textsc{StepBench}}
\newcommand{\eval}{\textsc{StepEval}}

\newcommand{\numrawarticles}{3{,}637}    
\newcommand{\nummethods}{10{,}305}       
\newcommand{\numtruetasks}{6{,}944}      
\newcommand{\numsamples}{6{,}885}        
\newcommand{\numsteps}{30{,}975}         
\newcommand{\numcats}{81}                
\newcommand{\numbencharticles}{2{,}920}  
\newcommand{\numexpertpct}{96.1\%}       

\newcommand{\numtasks}{5}
\newcommand{\numruns}{14}                
\newcommand{\nummodels}{12}              

\newcommand{\numannotators}{3}
\newcommand{\numannotated}{200}

\newcommand{\Tlast}{T1}      
\newcommand{\Tgap}{T2}       
\newcommand{\Terr}{T3}       
\newcommand{\Tord}{T4}       
\newcommand{\Tnext}{T5}      

\newcommand{\vpr}{\textsc{vpr}}          
\newcommand{\ned}{\textsc{ned}}          

\newcommand{\bestTfour}{63.0}

\newcommand{\bestTfournine}{33.1}

\newcommand{\qwenfourTfour}{38.0}
\newcommand{\qwenthirtytwoTfour}{48.6}
\newcommand{\qwenfourTplan}{11.2}
\newcommand{\qwenthirtytwoTplan}{73.8}

\newcommand{\rhoOrderVpr}{0.96}
\newcommand{\rhoErrVpr}{0.93}
\newcommand{\rhoGapVpr}{0.85}

\newcommand{\numtwosix}{214}
\newcommand{\numtwosixarticles}{94}
\newcommand{\numtwosixsteps}{1{,}034}
\newcommand{\numsharedcats}{27}

\newcommand{\deltaTone}{-1.4}
\newcommand{\deltaTgap}{-0.5}
\newcommand{\deltaTerr}{-0.6}
\newcommand{\deltaTord}{-8.3}
\newcommand{\deltaTnext}{-7.5}
\newcommand{\deltaTplan}{-16.2}
\newcommand{\maxcompdelta}{1.4}
\newcommand{\rhorankTone}{0.99}

\newcommand{\rhorankmin}{0.77}

\newcommand{\contamqwenTone}{-8.8}
\newcommand{\contamqwenthinkTone}{-2.3}

\newcommand{\goalwordsmain}{10.6}
\newcommand{\goalwordstwosix}{7.1}
\newcommand{\mainshortgoals}{73}
\newcommand{\twosixshortgoals}{99}

\newcommand{\goaldeltaTplan}{-19.7}

\newcommand{\thinkgapTord}{+1.0}
\newcommand{\thinkgapTplan}{+1.9}
\newcommand{\cmdthinkTord}{+7.4}
\newcommand{\cmdthinkTplan}{+14.6}

\title{\bench{}: Diagnosing Real-World Procedural Understanding \\in AI Agents}

\newcommand{\hficon}{\raisebox{-0.15em}{\includegraphics[height=1em]{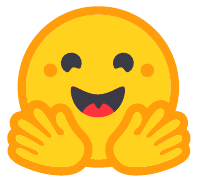}}}
\newcommand{\ghicon}{\raisebox{-0.15em}{\includegraphics[height=1em]{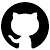}}}

  \author{
  \textbf{Hossein A.~Rahmani\textsuperscript{1}}\thanks{Corresponding author: \textit{\textcolor{sbPrimary}{hossein.rahmani.22@ucl.ac.uk}}}~
  \textbf{Bin Wu\textsuperscript{1}}~
  \textbf{Chaeeun Lee\textsuperscript{2}}~
  \textbf{Fanghua Ye\textsuperscript{1}}~
  \textbf{Sahan Bulathwela\textsuperscript{1}} \\
  \textbf{Sujay Kumar Jauhar\textsuperscript{3}}~
  \textbf{Marzieh Fadaee\textsuperscript{4}}~
  \textbf{Nick Craswell\textsuperscript{3}}~
  \textbf{Emine Yilmaz\textsuperscript{1}}
  \\[5pt]
  \textsuperscript{1}UCL AI Centre, University College London\\
  \textsuperscript{2}School of Informatics, University of Edinburgh\\
  \textsuperscript{3}Microsoft\\
  \textsuperscript{4}Cohere Labs
  \\[5pt]
  {\ghicon~\href{https://github.com/rahmanidashti/StepBench}{\texttt{StepBench}}}
  \quad
  {\hficon~\href{https://huggingface.co/datasets/rahmanidashti/StepBench}{\texttt{StepBench}}}
  \quad
  {\hficon~\href{https://huggingface.co/rahmanidashti/StepEval}{\texttt{StepEval}}}
}

\begin{document}
\maketitle
\begin{abstract}
Evaluating an agent's planning ability remains challenging for open-domain tasks, where verifying procedures through execution is often impractical.
Existing evaluations typically assess only the final plan and do not distinguish procedural errors such as missing, incorrect, or misordered actions.
We introduce \bench{}, a benchmark of \numsamples{} verified procedures crawled from wikiHow across \numcats{} open-domain categories.
From each procedure, we derive \numtasks{} diagnostic tasks: four with
deterministic answers involving withheld, incorrect, or shuffled steps,
and one requiring next-step generation. We additionally evaluate
full-plan generation to examine how diagnostic performance correlates
with end-to-end plan validity.
Using the same procedures across tasks lets us directly compare different planning capabilities.
For generation, \eval{} identifies critical failures that prevent goal completion.
Across \numruns{} configurations of \nummodels{} models, we find substantial differences across planning capabilities, with step ordering proving the most difficult: the strongest model correctly orders only \bestTfour{}\% of procedures, dropping to \bestTfournine{}\% for those with nine or more steps.
Incorrect-step detection also predicts judged plan validity across models, suggesting a lower-cost alternative for evaluating generation quality.
On \bench{}-Val, comprising \numtwosix{} procedures published in 2026, comprehension scores and rankings remain stable, suggesting limited influence from memorisation.
%
%
\end{abstract}

\section{Introduction} \label{sec:intro}
Large Language Model (LLM)-based agents are on behalf of humans to perform long-horizon tasks, with access to tools and actions that go beyond text generation.
These tools and actuators that can link AI systems beyond cyber-digital spaces will emerge, enabling LLM agents to tackle open-domain cyberphysical tasks (tasks that require manipulating both digital and physical environments).
AI agents capable of goal-based planning and tool calling \citep{shen2024taskbench,huang2024ultratool} on physical, open-domain tasks lead to synergistic teaming with humans (who have access to physical environments) at present and, in time, take autonomy over executing and completing these tasks.
Having a deep understanding of task decomposition, sequencing and gap identification in open-domain tasks plays a foundational role in creating reliable agents that can succeed in both knowledge-based and (much broader) cyber-physical tasks.
Open-domain task-planning benchmarks primarily focus on generating task decompositions \citep{zhang2021decompose,shen2024taskbench,huang2024ultratool} or predicting inter-step temporal dependencies \citep{zhang2020reasoning,yuan2023tasklama,lal2024cat}.
Fine-grained error analysis exists in formal \citep{valmeekam2023planbench,DBLP:conf/aaai/Kokel0SS25,DBLP:conf/iclr/KokelKSS26} and embodied \citep{DBLP:conf/nips/LiZWWZSGLLZLL0M24} environments, where an executor supplies ground truth, but is harder to obtain for open-domain written procedures.

%

%
Our goal is to \emph{diagnose} the capabilities underlying open-domain agent planning without executing the proposed steps. For open-domain tasks, generated plans generally cannot be physically executed or formally verified at benchmark scale, while evaluating only the final plan or task outcome provides little insight into which planning capabilities models struggle with. We aim to identify these strengths and weaknesses through targeted diagnostic tasks and examine how they relate to end-to-end plan generation performance.

\begin{figure*}[t]
    \centering
    \includegraphics[width=0.9\linewidth]{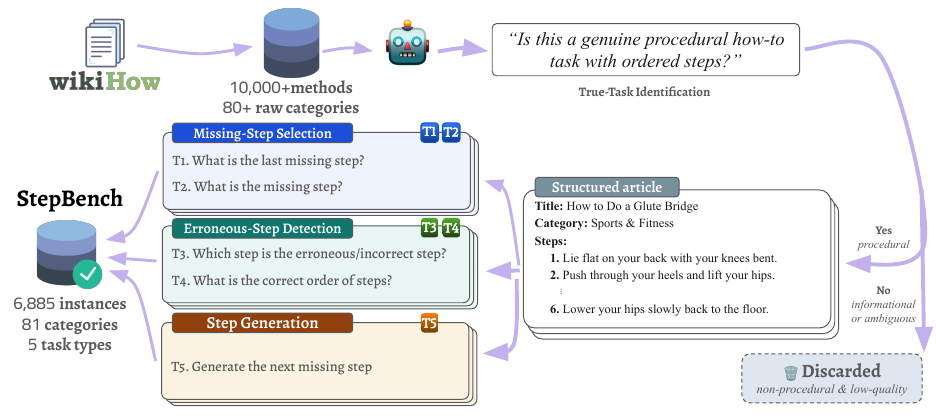}
    \caption{Constructing \bench{}. wikiHow methods are crawled into an initial pool, filtered by GPT-4.1 for whether they describe a genuine procedural task, and the selected procedures (each a task query, category, and ordered steps) instantiate five diagnostic tasks: missing-step selection (\Tlast{}--\Tgap{}), erroneous-step detection and reordering (\Terr{}--\Tord{}), and next-step generation (\Tnext{}). Full-plan generation is additionally evaluated to examine how diagnostic performance correlates with end-to-end plan validity.}
    \label{fig:pipeline}
\end{figure*}

We introduce \bench{}, comprising \numtasks{} tasks derived from \numsamples{} curated wikiHow procedures across \numcats{} categories.
\bench{}'s 5 diagnostic tasks probe distinct aspects of procedural understanding: last-step selection, missing-step recovery, incorrect-step detection, reordering, and next-step generation. The source articles have documented editorial provenance, with \numexpertpct{} co-authored by subject experts, and distractors are retrieved from other procedures to create plausible errors. Using the same procedures across tasks enables direct comparison of different aspects of agentic planning ability. For next-step  and full-plan generation, \eval{} adopts the critical-failure formulation of \citet{how2everything2026}, judging whether an output contains an error severe enough to prevent goal completion while allowing valid alternative plans.

Across \numruns{} configurations of \nummodels{} models, we find that performance varies substantially across capabilities. Models with similar incorrect-step detection scores can differ markedly in reordering. Even when all correct steps are provided, the strongest configuration recovers the reference order for only \bestTfour{}\% of procedures, falling to \bestTfournine{}\% for procedures with nine or more steps. Reordering accuracy also correlates strongly with judged full-plan validity ($\rho=\rhoOrderVpr{}$), suggesting that diagnostic tasks can provide a lower-cost signal of plan quality. 

Our main contributions are as follows:

\begin{itemize}[leftmargin=1.4em,itemsep=1pt,topsep=3pt]
  \item We introduce \bench{}, a planning diagnosis benchmark suite for evaluating plan recovery, error detection, ordering, and generation on \numsamples{} open-domain procedures. We additionally evaluate on \numtwosix{} procedures published in 2026 (\S\ref{sec:data}--\S\ref{sec:benchmark}).
  \item We apply critical-failure evaluation to next-step and full-plan generation through \eval{}, an 8B judge distilled from GPT-5 and validated against human annotations (\S\ref{sec:stepeval}).
  \item We evaluate \nummodels{} models across \numruns{} configurations and find substantial differences across planning abilities. Diagnostic performance also correlates with judged full-plan validity, providing a lower-cost signal of generation quality (\S\ref{sec:results}).
\end{itemize}
\section{Related Work}
\label{sec:related}

\paragraph{Task Decomposition for Planning.}
Language models can decompose high-level goals into executable steps, evaluated against task-management logs and wikiHow \citep{zhang2021decompose}, represented as task graphs \citep{yuan2023tasklama}, or incorporated into tool-use pipelines \citep{shen2024taskbench,huang2024ultratool}. Evaluating only the final plan, however, can obscure distinct failure modes: necessary actions may be missing, incorrect actions introduced, or correct actions placed in the wrong order. Several benchmarks separate such planning abilities when a simulator, formal domain, or annotated workflow provides an oracle \citep{DBLP:conf/nips/LiZWWZSGLLZLL0M24,DBLP:conf/aaai/Kokel0SS25,DBLP:conf/iclr/KokelKSS26,DBLP:conf/iclr/QiaoFQW0JXHC25}. Embodied Agent Interface, for example, executes model-generated action sequences in household simulators and uses the resulting state to label failures as missing, additional, or misordered steps \citep{DBLP:conf/nips/LiZWWZSGLLZLL0M24}, while ACPBench uses planning domain definition language \citep{DBLP:journals/aim/McDermott00} to automatically generate planning questions with verifiable ground-truth answers, covering capabilities such as action applicability and plan validity \citep{DBLP:conf/aaai/Kokel0SS25}. \bench{} instead evaluates open-domain procedures that cannot be checked against a simulator or formal model. Its tasks are derived from the same underlying procedures, enabling direct comparisons between component procedural skills and full-plan generation.

\paragraph{Procedural Reasoning and Plan Verification.}
Related benchmarks capture individual components of this profile. \citet{zhang2020reasoning} introduce wikiHow goal--step and pairwise temporal-ordering tasks with semantically similar negatives, while other procedural benchmarks test temporal relations or step ordering \citep{lal2024cat,anika2025ordering,sakaguchi2021proscript}. Formal planning benchmarks verify complete plans through execution \citep{valmeekam2022large,valmeekam2023planbench}, and reasoning benchmarks localise errors in mathematical solutions or long chains of thought \citep{DBLP:conf/acl/ZhengZZLLYLZL25,DBLP:conf/acl/HeLLWB0P0ZS025}. \bench{} brings these abilities together for open-domain procedures, relating error localisation and full-order recovery to generation for the same goals. This joint evaluation is relevant to agents that must audit as well as generate plans, consistent with prior findings that language models can struggle to identify errors in their own outputs \citep{DBLP:journals/corr/abs-2310-08118,DBLP:conf/iclr/StechlyVK25}.

\paragraph{Evaluating Generated Plans.}
Open-ended plan evaluation has typically relied on surface overlap \citep{zhang2021decompose,shen2024taskbench,brahman2024plasma}, with recent benchmarks increasingly using LLM judges \citep{huang2024ultratool,how2everything2026}. To evaluate next-step and full-plan generation, we follow the critical-failure formulation of \citet{how2everything2026} and distil an 8B judge from GPT-5's judgments, validating it against human annotations (\S\ref{sec:stepeval}). Unlike evaluations focused solely on final generated plans, \bench{} probes distinct components of procedural planning through targeted \emph{diagnostic} tasks. Appendix~\ref{app:related} provides a detailed benchmark comparison and further discussion of procedural corpora and LLM-judge reliability.
\section{\bench{}: Data Curation}
\label{sec:data}

\bench{} is built from wikiHow procedures that pass a two-stage filtering and curation pipeline (see Figure~\ref{fig:pipeline}). We crawl \numrawarticles{} wikiHow articles across \numcats{} categories. wikiHow suits this purpose because its articles are written as explicitly ordered imperative steps, span diverse open-domain tasks, and carry an inspectable editorial history. \numexpertpct{} of articles behind the final benchmark are co-authored by a credentialed subject expert, and a substantial fraction has additionally been fact-checked, expert-approved, or physically tested. Further details are provided in Table~\ref{tab:corpus} and Figure~\ref{fig:corpus} in Appendix \ref{app:data}. While these features do not prove every procedure correct, they are evidence that the procedures were written for real users and reviewed through an accountable process.

\label{sec:data-pipeline}
wikiHow articles often contain multiple \emph{task queries} and \emph{methods}, each with its own ordered sequence of steps. We treat each method as an independent procedure, yielding \nummethods{} candidate procedures from the initial crawl. To ensure procedural quality, we use GPT-4.1 \citep{openai2025gpt41} to filter and refine candidates in two stages (prompts are included in Appendix~\ref{app:pipeline-prompts}). \textbf{(1)~Task filtering} retains only candidates that describe a well-defined procedural task with a specific objective achieved through an ordered sequence of steps. We exclude, for example, collections of advice or reflective prompts presented in a step-like format. This leaves \numtruetasks{} of the original \nummethods{} candidates. \textbf{(2)~Step curation} applies two checks to each remaining procedure: a \emph{necessity} check removes steps not required to complete the task (e.g., encouragements, asides, and optional variations), while an \emph{order} check reorders the remaining steps where necessary to form an executable sequence. This yields \numsamples{} curated procedures, each represented by its category, task query, and ordered reference steps. Only 59 candidates are discarded during step curation. This limited attrition may partly reflect the expert co-authorship and editorial review of the source articles described above. All tasks draw from this shared pool, enabling direct cross-task comparison (\S\ref{sec:results-proxy}).

\paragraph{Human validation.}
\label{sec:data-human-eval}
We assess both task filtering and step curation validity on \numannotated{} instances each, with \numannotators{} independent annotators per instance. Filtering decisions agree with the human majority on 82.0\% of instances, and 81.5\% of curated procedures are judged correct by a majority. Sampling protocol, instructions, and inter-annotator agreement are in Appendix~\ref{app:data-human-eval}.

\paragraph{\bench{}-Val.}
\label{sec:data-2026}
The main corpus was crawled by 2025-04-10, before the release of some evaluated models, raising the risk of training-data contamination. To account for this, we construct a separate validation set from wikiHow articles \emph{created in} 2026, after the release of the 11 evaluated models (\S\ref{sec:setup}). We also evaluate GPT-5.6-sol on this set, but do not establish that these articles postdate its training or release. Applying the same two-stage pipeline yields \numtwosix{} procedures (\numtwosixsteps{} steps from \numtwosixarticles{} articles), spanning \numsharedcats{} of the \numcats{} categories. This set is excluded from the main \bench{} benchmark and its headline results. We report the corresponding evaluation results in Appendix~\ref{app:contam}.

\section{The \bench{} Task Suite}
\label{sec:benchmark}

Agents can fail at procedural tasks in different ways: identifying the final step, detecting missing or incorrect steps, or recovering the correct order. These require distinct capabilities that no single task can fully assess. We therefore define five complementary diagnostic tasks over the same \numsamples{} verified procedures, which together constitute our benchmark. By evaluating these tasks on a shared set of procedures, we can attribute performance differences to the capabilities being tested rather than variations in the underlying data. We further examine how performance on these diagnostic tasks correlates with full plan generation, assessing whether they serve as effective indicators of agents' procedural capabilities. Figure~\ref{fig:stepbench} in Appendix~\ref{app:data} illustrates the five diagnostic tasks and the procedure generation task using a single example.

\subsection{Task Definitions}
\label{sec:benchmark-tasks}

\paragraph{\Tlast{}: Last-step selection.}
Given the task query, all steps except the last, and four candidate final steps, the model selects the correct one. 

\paragraph{\Tgap{}: Missing-step recovery.}
Similar to \Tlast{}, but the withheld step can occur at any position, with a gap marking its location in the sequence. The model must identify the missing step using both preceding and following context.

\paragraph{\Terr{}: Incorrect-step detection.}

A step at a random position is replaced with a retrieved in-category step (\S\ref{sec:benchmark-distractors}) from another procedure. The model must identify its index without knowing where the replacement occurred, testing its ability to verify a complete, superficially coherent plan against the task goal.

\paragraph{\Tord{}: Step reordering.}
The reference steps are shuffled and renumbered by their shuffled positions, and the model must recover the original ordering. Since all steps are present, this isolates ordering knowledge from content knowledge.

\paragraph{\Tnext{}: Next-step generation.}
Given the task goal and a partial sequence of steps, the model generates the next step without candidate options. We randomly select where to stop the sequence, with at least one preceding and one remaining step, covering early, middle, and late continuations. Unlike \Tgap{}, the model must generate the step without candidate options.

\paragraph{Full plan generation.}
Given only the task goal, the model generates the complete ordered procedure. We examine how performance on the diagnostic tasks correlates with full plan generation, assessing the extent to which the capabilities measured by our benchmark are predictive of procedural generation performance.

\subsection{Distractor Retrieval}
\label{sec:benchmark-distractors}
\Tlast{}, \Tgap{}, and \Terr{} require plausible distractor steps, which we retrieve from other procedures in \bench{}. We embed all steps using Qwen3-Embedding-8B \citep{zhang2025qwen3embeddingadvancingtext} and rank candidates by cosine similarity. For \Terr{}, we replace one reference step with an in-category distractor to prevent detection based on topic similarity alone. For \Tlast{} and \Tgap{}, we pair the correct step with three distractors, one from each of the following strategies:

\begin{itemize}[leftmargin=1.4em,itemsep=1pt,topsep=3pt]
  \item \textbf{Step-similar.} The most similar step to the withheld step, drawn from another task. These are near-misses involving the same action at a different granularity or with different parameters, making them the hardest distractors and testing procedural detail rather than topic alone.
  \item \textbf{In-category.} The step most similar to the procedure as a whole, drawn from another procedure in the same category. The distractor is topically relevant but belongs to a different task.
  \item \textbf{Out-of-category.} The same retrieval strategy, but restricted to a different category. These are easier to reject and serve as a basic check that the model is engaging with the task.
\end{itemize}

\section{\eval{}: Detecting Critical Failures}
\label{sec:stepeval}

\begin{table*}[t]
\centering
\caption{
Examples of critical-failure evaluation with \eval{} on full-plan generation. Each example pairs a task query and reference steps with a plan generated by Llama~3.1~8B Instruct and the judgment of the \eval{}. The examples show an omitted essential action, failure to satisfy the task goal, and an acceptable variation. Steps and explanations are abbreviated for readability, and bold text highlights the relevant differences.
}
\label{tab:stepeval-examples}

\begingroup
\footnotesize
\setlength{\tabcolsep}{5pt}
\renewcommand{\arraystretch}{1.15}

\begin{tabular}{
    @{}
    >{\raggedright\arraybackslash}p{0.39\textwidth}
    >{\raggedright\arraybackslash}p{0.35\textwidth}
    >{\raggedright\arraybackslash}
        p{\dimexpr0.26\textwidth-4\tabcolsep\relax}
    @{}
}
\toprule
\rowcolor{sbPale}
\textbf{Task query and reference steps}
&
\textbf{Model-generated steps}
&
\textbf{\eval{}}
\\
\midrule

\textbf{Query:}
\textcolor{sbMuted}{[Cakes]}
How to make cake flour from all-purpose flour?
\par\smallskip
\textbf{Reference:}
\par
\textbf{1.} Measure one level cup of all-purpose flour.
\textbf{2.} \textcolor{sbAccentB}{\textbf{Remove two tablespoons of flour.}}
\textbf{3.} \textcolor{sbAccentB}{\textbf{Replace them with two tablespoons of cornstarch.}}
\textbf{4.} Sift the flour and cornstarch together five times.
&
\textbf{1.} Measure one cup of all-purpose flour.
\textbf{2.} Sift into a bowl to remove lumps.
\textbf{3.} Sift again through a fine-mesh sieve.
\textbf{4.} Repeat sifting two or three more times until powdery and smooth.
\textbf{5.} Store the cake flour in an airtight container.
&
\textcolor{sbAccentB}{\textbf{Missing critical action.}}
\par\smallskip
The plan omits the cornstarch substitution. Repeated sifting
leaves the ingredient composition unchanged and does not
achieve the conversion described by the reference.
\\
\midrule

\textbf{Query:}
\textcolor{sbMuted}{[Computers and Electronics]}
How to blur your background in Zoom using the mobile app?
\par\smallskip
\textbf{Reference:}
\par
\textbf{1.} Open the Zoom app.
\textbf{2.} Join a meeting.
\textbf{3.} Tap the three-dot menu.
\textbf{4.} Tap Virtual Background or Background and Filters.
\textbf{5.}
\textcolor{sbAccentB}{\textbf{Tap Blur.}}
&
\textbf{1.} Open the Zoom mobile app.
\textbf{2.} Join or start a meeting.
\textbf{3.} Tap More or the three-dot icon.
\textbf{4.} Select Virtual Background.
\textbf{5.}
\textcolor{sbAccentB}{\textbf{Choose an available background or upload your own.}}
\textbf{6.} Select the background to apply it.
\textbf{7.} Adjust it through the Background option as needed.
&
\textcolor{sbAccentB}{\textbf{Failure to achieve the stated goal.}}
\par\smallskip
The plan reaches the relevant menu but instructs the user to
apply a replacement background, omitting the essential action
of selecting Blur.
\\
\midrule

\textbf{Query:}
\textcolor{sbMuted}{[Computers]} How to check your computer uptime on Linux?
\par\smallskip
\textbf{Reference:}
\par
\textbf{1.} Open the terminal.
\textbf{2.} Type
\textcolor{sbPrimary}{\textbf{\texttt{uptime -p}}}
and press Enter.
&
\textbf{1.} Open the terminal.
\textbf{2.} Type \textcolor{sbPrimary}{\textbf{\texttt{uptime}}} and press Enter.
\textbf{3.} Read the displayed system uptime.
&
\textcolor{sbPrimary}{\textbf{No critical failure.}}
\par\smallskip
Omitting \texttt{-p} changes the output format, but the
command still provides the requested uptime information.
\\

\bottomrule
\end{tabular}
\endgroup
\end{table*}

Four of the five diagnostic tasks (\Tlast{}--\Tord{}) have known ground-truth answers and are evaluated using accuracy, with \Tord{} additionally reporting normalised edit distance. Evaluating next-step (\Tnext{}) and full-plan generation is less straightforward: valid procedures may differ substantially from the reference, making text-overlap metrics insufficient, while human evaluation is costly and difficult to scale \citep{brahman2024plasma}. We use \eval{} valid-plan rate (\vpr{}; \S\ref{sec:stepeval-protocol}) as our primary metric, and additionally report ROUGE \citep{lin2004rouge} and baseline-rescaled BERTScore \citep{zhang2020bertscore}. Following the critical-failure formulation of \citet{how2everything2026}, \eval{} assesses whether a generated procedure contains any failure severe enough to prevent task completion or render the instructions unusable. We also validate the judge's per-instance labels against human annotations and distil a lower-cost judge for large-scale evaluation.

\subsection{\eval{} Protocol}
\label{sec:stepeval-protocol}
Following the formulation of \citet{how2everything2026}, we define a \emph{critical failure} as an omission, extraneous or counterproductive action, contradiction, severe vagueness, or other error that prevents task completion or renders the instructions unusable. Given the task query, reference steps, and generated plan, failures are classified into three families: \emph{contradictions} with the goal or reference steps; \emph{structural faults}, including internal inconsistency, incoherence, and severe vagueness; and \emph{missing or extraneous actions}, such as omitted essential steps, counterproductive additions, or unwarranted repetition. The reference steps help establish essential actions and their dependencies, but differences in wording, granularity, emphasis, or ordering, as well as harmless additions, are accepted if the necessary actions are preserved. The judge returns free-text reasoning and a list of failures, each linked to the relevant reference and candidate steps. Full guidelines and prompts are provided in Appendix~\ref{app:stepeval}.

We derive a binary failure label from the judge's output and compute the \textbf{valid-plan rate} (\vpr{}) as the proportion of plans with no critical failures:

\[
  \vpr{}(D) \;=\; \frac{1}{|D|}\sum_{x \in D}
    \mathbb{I}\!\left[\,J(g, S^\star, \hat{S}) = \textsf{no\_failure}\,\right]
\]

where $g$ is the task goal, $S^\star$ the reference procedure, $\hat{S}$ the generated plan, and $J$ the judge. Table \ref{tab:stepeval-examples} shows representative critical failure examples.

\begin{table*}[t]
\centering\small
\setlength{\tabcolsep}{4.2pt}
\adjustbox{max width=\textwidth}{%
\begin{tabular}{l r r r r r r r}
\toprule
& \multicolumn{6}{c}{\textbf{Diagnostic tasks}} & \textbf{Full plan} \\
\cmidrule(lr){2-7}\cmidrule(lr){8-8}
\textbf{Model} & \Tlast{} Acc. & \Tgap{} Acc. & \Terr{} Acc. & \Tord{} Acc. & \Tord{} \ned{} & \Tnext{} \vpr{} & \multicolumn{1}{c}{\vpr{}} \\
\midrule
\headrow \multicolumn{8}{c}{\textit{Proprietary}} \\
\midrule
GPT-5.6-sol & 75.7 & 73.6 & 66.5 & \best{63.0} & \best{80.6} & \best{79.7} & \best{92.1} \\
Gemini 3 Flash \textsuperscript{T=0} & \best{78.8} & \best{78.2} & \best{71.1} & 58.7 & 73.4 & 77.0 & \second{84.5} \\
Gemini 3 Flash \textsuperscript{T=1} & \second{77.4} & \second{77.1} & \second{69.8} & \second{59.4} & \second{76.6} & \second{77.7} & 83.0 \\
Command A Reasoning & 67.9 & 43.9 & 60.5 & 51.8 & 74.1 & 70.0 & 82.0 \\
Command A & 65.5 & 59.1 & 61.2 & 44.3 & 70.0 & 70.7 & 67.4 \\
\midrule
\headrow \multicolumn{8}{c}{\textit{Open weight}} \\
\midrule
Qwen3 32B \textsuperscript{\scriptsize think} & 65.4 & 43.6 & 58.4 & 49.6 & 72.4 & 73.3 & 75.7 \\
Qwen3 32B & 62.7 & 30.5 & 57.3 & 48.6 & 72.4 & 69.7 & 73.8 \\
Gemma 3 27B & 62.0 & 47.2 & 57.4 & 47.1 & 71.5 & 68.3 & 73.5 \\
Qwen3 14B & 53.9 & 24.9 & 56.9 & 45.3 & 70.3 & 71.5 & 66.2 \\
GLM-4 9B & 54.3 & 28.3 & 53.9 & 38.3 & 65.5 & 65.9 & 60.6 \\
Qwen3 8B & 28.2 & 23.5 & 51.5 & 39.8 & 66.5 & 67.4 & 55.0 \\
Llama 3.1 8B & 21.6 & 26.3 & 48.0 & 16.0 & 45.9 & 50.4 & 53.6 \\
Qwen3 4B Instruct \textsuperscript{\scriptsize 2507} & 42.4 & 28.2 & 49.9 & 35.2 & 62.8 & 68.2 & 57.3 \\
Qwen3 4B & 25.4 & 28.2 & 49.7 & 38.0 & 65.2 & 64.1 & 11.2 \\
\midrule
\textit{Random baseline} & 25.0 & 25.0 & \multicolumn{1}{c}{--} & \multicolumn{1}{c}{--} & \multicolumn{1}{c}{--} & \multicolumn{1}{c}{--} & \multicolumn{1}{c}{--} \\
\bottomrule
\end{tabular}}
\caption{Performance on the five diagnostic tasks and end-to-end full-plan generation (\%). \Tlast{}--\Tord{} report exact-match accuracy. \Tord{} additionally reports normalised edit distance (\ned{}), with higher values indicating closer agreement with the reference order. \Tnext{} and full-plan generation are evaluated using \eval{} valid-plan rate (\vpr{}). Best scores within each column are \best{bold}; second-best scores are \second{underlined}.}
\label{tab:main}
\end{table*}

\paragraph{Human validation.}
We validate \eval{} on 200 generated plans assessed by three human annotators, who identify critical failures using the task query, reference steps, and generated plan. GPT-5 agrees with the human majority on 85.5\% of examples, including 87.1\% of \texttt{has\_failure} and 83.8\% of \texttt{no\_failure} cases. Annotation details and inter-annotator agreement are reported in Appendix~\ref{app:stepeval-human-eval}.

\subsection{Cost-Efficient Planning Judge at Scale}
\label{sec:stepeval-distill}

GPT-5 \citep{openai2026gpt5} achieves 85.5\% agreement with human annotations, but its cost makes large-scale evaluation impractical. We therefore distil GPT-5's judgements into \eval{}, a smaller Qwen3-8B judge designed for cost-efficient evaluation. We collect 82K GPT-5 annotations on plans generated by diverse models and fine-tune Qwen3-8B for three epochs. On the human-labelled evaluation set, \eval{} achieves 92\% agreement with GPT-5 and 82.5\% with the human majority, including 80.2\% on \texttt{has\_failure} and 84.8\% on \texttt{no\_failure} cases. These results suggest that the distilled judge retains strong agreement across both labels while reducing evaluation costs. Further training details are provided in Appendix~\ref{app:distill}.

\section{Experimental Setup}
\label{sec:setup}

\paragraph{Models.}
We evaluate \nummodels{} open-weight and proprietary models spanning an order of magnitude in parameter size. Evaluated models include Qwen3-4B, Qwen3-8B, Qwen3-14B, Qwen3-32B, and Qwen3-4B-Instruct-2507 \citep{qwen3}, Llama-3.1-8B-Instruct \citep{dubey2024llama3}, Gemma-3-27B-IT \citep{gemma2024}, and GLM-4-9B \citep{glm2024}, as well as proprietary models GPT-5.6 Sol, Gemini~3 Flash \citep{gemini2025}, Command~A, and its reasoning-tuned variant Command~A~Reasoning \citep{cohere2025commanda}. All \numruns{} configurations have results on all diagnostic tasks and full-plan generation on both datasets. We use the fine-tuned 8B \eval{} judge for next-step and full-plan VPR throughout the results. GPT-5 is used only as a reference in judge training and validation. Further model details are provided in Appendix~\ref{app:experiment}.

\paragraph{Evaluation.}
We evaluate each configuration on both \bench{} and \bench{}-Val (\S\ref{sec:data-2026}), using identical prompts, decoding settings, parsers, and judges. First four diagnostic tasks (\Tlast{}--\Tord{}) are scored by exact match, with normalised edit distance additionally reported for \Tord{}. For next-step (\Tnext{}) and full-plan generation, we use \eval{} valid-plan rate (\S\ref{sec:stepeval-protocol}) as our primary metric, and additionally report ROUGE-1/2/L and baseline-rescaled BERTScore.

\section{Results and Discussion}
\label{sec:results}

\begin{figure*}[t]
    \centering
    \includegraphics[width=0.92\textwidth]{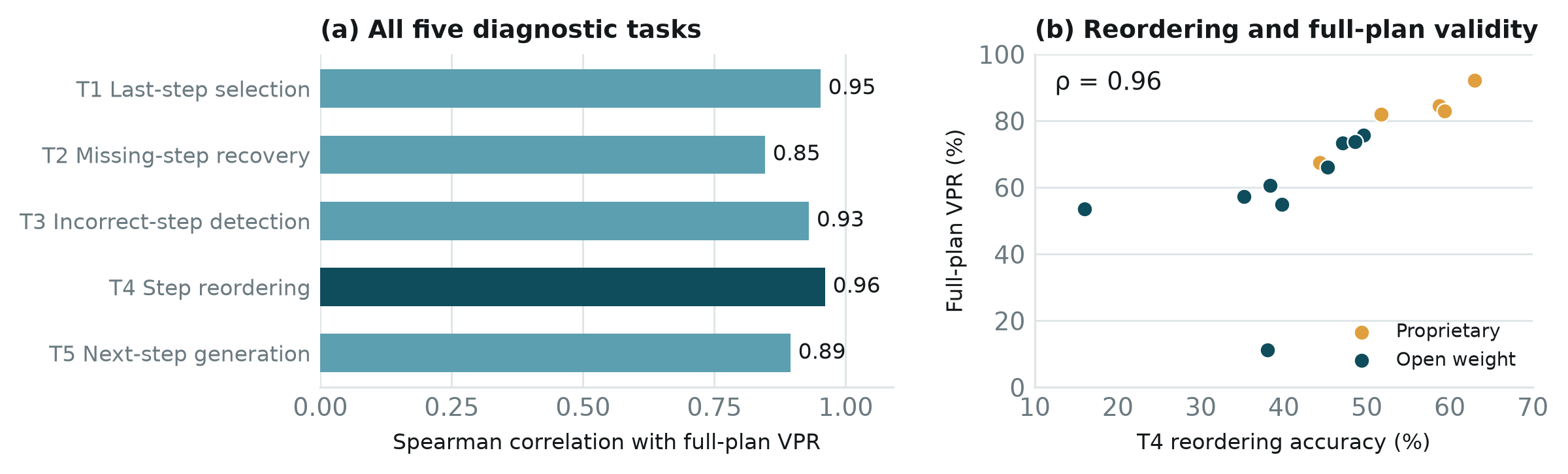}
    \caption{Correlations between the five diagnostic tasks and end-to-end plan validity. Left: Spearman correlations with full-plan VPR across all \numruns{} configurations, using accuracy for \Tlast{}--\Tord{} and VPR for \Tnext{}. Right: reordering accuracy versus full-plan VPR, with each point representing one configuration.}
    \label{fig:proxy}
\end{figure*}

We examine what StepBench's five diagnostic tasks reveal about planning capabilities and how they relate to full-plan generation. Table~\ref{tab:main} reports results across all \numruns{} configurations. Despite being substantially less costly to evaluate than full plans, with \Tlast{}--\Tord{} requiring no judge model at all, the diagnostic scores are strongly associated with full-plan validity while also exposing capability differences that full-plan evaluation alone can obscure.

\subsection{Planning Beyond Aggregate Scores}
\label{sec:results-separable}

Table~\ref{tab:main} shows that similar full-plan validity can hide differences in specific planning capabilities. Command~A and Qwen3~14B have similar full-plan VPR (67.4\% and 66.2\%), but differ in missing-step recovery (59.1\% and 24.9\%). Their reordering accuracy is much closer (44.3\% and 45.3\%). The diagnostic tasks therefore reveal a difference in recovering missing content that full-plan VPR alone does not show.

Success on one diagnostic task also does not imply success on the others. Qwen3~4B and its Instruct-2507 variant have nearly identical incorrect-step detection accuracy (49.7\% and 49.9\%), but different full-plan VPR (11.2\% and 57.3\%). The original model has slightly higher reordering accuracy (38.0\% versus 35.2\%). These results support evaluating recovery, detection, ordering, and continuation separately. Each task tests a specific capability, while full-plan generation requires producing both the actions and their sequence.

\subsection{Correlation with Full-Plan Validity}
\label{sec:results-proxy}
Section~\ref{sec:results-separable} shows that the diagnostics expose capability differences that full-plan VPR hides. Figure~\ref{fig:proxy} shows how each diagnostic task relates to full-plan VPR across all \numruns{} configurations. All five tasks are positively associated with full-plan validity. Reordering has the strongest correlation ($\rho=\rhoOrderVpr{}$), followed by last-step selection ($\rho=0.95$), incorrect-step detection ($\rho=\rhoErrVpr{}$), next-step VPR ($\rho=0.89$), and missing-step recovery ($\rho=\rhoGapVpr{}$). Thus, models that perform well on the diagnostic tasks also tend to generate more valid full plans over the same procedure pool.

These associations suggest that the diagnostic tasks provide a lightweight signal of planning ability without requiring full-plan generation and costly judge evaluation. In particular, \Tlast{}--\Tord{} have known answers and require no judge model. The correlations should nevertheless be interpreted at the model level: strong performance on a diagnostic task does not imply valid full-plan generation for every procedure on which the model succeeds in that task. Taken together, the diagnostics complement full-plan evaluation by capturing overall planning performance while revealing capability differences that full-plan VPR can obscure.

\subsection{Procedure Length Effects}
Table~\ref{tab:main} shows that the best reordering accuracy is \bestTfour{}\%, even though all required steps are supplied. This is below the best accuracy on selection, recovery, and detection. Figure~\ref{fig:length} in Appendix~\ref{app:length} shows how performance changes with procedure length. For GPT-5.6-sol, exact-order accuracy falls from 86.8\% at two steps to 33.1\% at nine or more steps, while normalised edit distance falls from 86.8 to 76.6. The model often produces an order close to the reference, but recovering the complete sequence becomes harder as the length increases. Exact reference agreement is a strict diagnostic: a different order is not necessarily an invalid plan when some actions can be exchanged.

Length affects detection differently. Gemini~3 Flash's incorrect-step detection accuracy rises from 68.5\% at four steps to 80.3\% at nine or more, whereas Qwen3~4B falls from 51.1\% at three steps to 42.5\% at seven or eight. Longer procedures therefore do not make every diagnostic task harder in the same way, and aggregate scores can hide these differences.

\subsection{Effects of Scale and Reasoning}
\label{sec:results-scale}

Table~\ref{tab:main} shows that increasing model size changes some capabilities more than others. From Qwen3~4B to Qwen3~32B without thinking, reordering accuracy rises from \qwenfourTfour{}\% to \qwenthirtytwoTfour{}\%, while full-plan VPR rises from \qwenfourTplan{}\% to \qwenthirtytwoTplan{}\%. Incorrect-step detection changes less, from 49.7\% to 57.3\%.

Reasoning comparisons also show uneven changes. Qwen3~32B gains \thinkgapTord{} points on reordering and \thinkgapTplan{} on full-plan VPR with thinking enabled. Command~A~Reasoning improves on Command~A by \cmdthinkTord{} and \cmdthinkTplan{} points, respectively. However, missing-step recovery improves by 13.2 points for Qwen3~32B and falls by 15.2 points for Command~A~Reasoning. These comparisons show why gains in full-plan generation should not be taken to imply gains in every diagnostic capability. The Command~A comparison also involves different model variants, so it does not isolate the effect of reasoning alone.

\subsection{Performance on \bench{}-Val}
\label{sec:results-contam}

Table~\ref{tab:contam} shows performance on \bench{}-Val after matching category composition and procedure length. Selection, recovery, and detection change by at most \maxcompdelta{} points on average, while reordering and next-step VPR decrease by 8.3 and 7.5 points, respectively. Full-plan VPR decreases by 16.2 points. Overall, the diagnostic tasks show varying sensitivity to the dataset change, while remaining comparatively stable.

Model rankings are also strongly preserved across datasets, with rank correlations ranging from \rhorankmin{} to \rhorankTone{}. This consistency suggests that the diagnostic patterns observed on \bench{}, although it does not by itself rule out memorisation because the sets differ in construction and GPT-5.6-sol's training chronology is unverified. Appendix~\ref{app:contam} provides per-configuration comparisons and additional controls.

\begin{table}[t]
\centering\small
\setlength{\tabcolsep}{4pt}
\begin{tabular}{l r r r r r r}
\toprule
\headrow & \Tlast{} & \Tgap{} & \Terr{} & \Tord{} & \Tnext{} & Full plan \\
\midrule
Change & \deltaTone{} & \deltaTgap{} & \deltaTerr{} & \deltaTord{} & \deltaTnext{} & \deltaTplan{} \\
\bottomrule
\end{tabular}
\caption{Mean score change on \bench{}-Val relative to \bench{} (percentage points), after matching categories and reweighting procedure lengths. The five diagnostic tasks use accuracy (\Tlast{}--\Tord{}) or VPR (\Tnext{}), while full-plan generation uses VPR. Results are averaged over all \numruns{} configurations.}
\label{tab:contam}
\end{table}

\section{Conclusion}
\label{sec:conclusion}

We introduce \bench{}, a benchmark for diagnosing open-domain procedural planning through five complementary tasks over \numsamples{} curated procedures. Existing open-domain benchmarks primarily evaluate final generated
plans \citep{zhang2021decompose,shen2024taskbench,huang2024ultratool},
while fine-grained error analysis has largely focused on formal or
embodied settings with executable plans
\citep{valmeekam2023planbench,DBLP:conf/aaai/Kokel0SS25,DBLP:conf/nips/LiZWWZSGLLZLL0M24}. \bench{} bridges this gap by probing distinct aspects of procedural understanding, enabling fine-grained diagnosis of planning strengths and weaknesses without requiring execution.

Our evaluation of \nummodels{} models across \numruns{} configurations reveals substantial differences in diagnostic profiles. Reordering remains particularly challenging even when all correct steps are provided: the strongest configuration recovers the reference order for only \bestTfour{}\% of procedures, dropping to \bestTfournine{}\% for procedures with nine or more steps. Models with similar incorrect-step detection scores can also differ markedly in reordering, highlighting differences that aggregate planning scores may obscure.

All five diagnostic tasks are positively correlated with judged
end-to-end plan validity, with reordering showing the strongest
association ($\rho=\rhoOrderVpr{}$). These findings suggest that
diagnostic tasks can provide lower-cost indicators of planning performance
while revealing strengths and weaknesses that full-plan evaluation alone
cannot capture. Evaluation on \bench{}-Val, comprising \numtwosix{}
procedures published in 2026, further supports the robustness of these
findings, with relatively stable diagnostic performance and preserved
model rankings across older and newer procedures.

Overall, \bench{} provides a framework for fine-grained planning
diagnosis, connecting distinct procedural capabilities to end-to-end generation performance. We release the corpus, task suite, model outputs, judge annotations, and analysis code to support further research on diagnosing procedural planning in open-domain settings.

\section*{Limitations}
\bench{} focuses on English-language procedures from wikiHow, primarily covering everyday tasks. Broader coverage of languages, sources, and specialised domains would strengthen its generality. The procedures are also relatively short, leaving longer and more complex planning scenarios for future evaluation.

\eval{} provides a scalable assessment of critical failures, but its judgements depend on reference quality and the interpretation of acceptable procedural variation. Additional references and expert assessments could improve coverage of alternative valid plans. Computational resources and API credit budgets also limited the number of models and configurations we could evaluate; expanding this coverage would further strengthen the empirical findings.

\section*{Acknowledgments}
This research is supported by the Engineering and Physical Sciences Research Council [EP/S021566/1] and the EPSRC Fellowship titled ``Task Based Information Retrieval'' [EP/P024289/1]. This project also used the DGX Spark compute equipment provided by NVIDIA for the grant titled ``Agentic Evaluation: Efficient and Reliable Evaluation for Agentic AI Systems''. This research is supported in part by the Cohere Labs Catalyst Grant Program. 


\bibliography{references}

\appendix

\section{Extended Related Work}
\label{app:related}

\subsection{Benchmark Comparison}
Table~\ref{tab:related} positions \bench{} against selected benchmarks most closely related to task decomposition or procedural comprehension. It is a map of task coverage, rather than a ranking of resources or a claim that the benchmarks solve identical problems. In particular, the source column records where procedures originate; it is not a measure of reference correctness. The coverage columns distinguish plan production, recovery of a withheld step, localisation of an inserted incorrect step, and ordering. They make explicit that \bench{} combines diagnostic comprehension and generation over the same procedure instances. We report separate task scores because similar overall performance can conceal different procedural strengths and weaknesses. Prior work shows that a few underlying capabilities explain much of the variation in benchmark scores across models \citep{DBLP:conf/nips/RuanMH24}. \bench{} provides a more detailed procedural profile by evaluating missing-step recovery, incorrect-step detection, ordering, and generation over the same procedure instances.

\begin{table*}[t!]
\centering\small
\setlength{\tabcolsep}{3.6pt}
\begin{tabular}{l l r c c c c l}
\toprule
\textbf{Benchmark} & \textbf{Source} & \textbf{Size} & \textbf{Gen.} & \textbf{Miss.} & \textbf{Err.} & \textbf{Ord.} & \textbf{Decomp. metric} \\
\midrule
\multicolumn{8}{c}{\textit{Task decomposition}} \\
\midrule
\citet{zhang2021decompose} & Wunderlist, wikiHow & 8{,}262 & \ding{51} & \ding{55} & \ding{55} & $\circ$ & ROUGE, BERTScore \\
TaskLAMA \citep{yuan2023tasklama} & search logs, MSCT & 1{,}612 & \ding{51} & \ding{55} & \ding{55} & $\circ$ & ROUGE, Hungarian \\
TaskBench \citep{shen2024taskbench} & synth.\ tool graphs & 17{,}331 & \ding{51} & \ding{55} & \ding{55} & \ding{55} & ROUGE only \\
UltraTool \citep{huang2024ultratool} & expert queries & 5{,}824 & \ding{51} & \ding{55} & \ding{55} & \ding{55} & LLM judge (6 dim.) \\
\midrule
\multicolumn{8}{c}{\textit{Procedural comprehension}} \\
\midrule
CaT-Bench \citep{lal2024cat} & recipes & 4{,}260 & \ding{55} & \ding{55} & \ding{55} & $\circ$ & accuracy, F1 \\
\citet{anika2025ordering} & recipes & --- & \ding{55} & \ding{55} & \ding{55} & \ding{51} & Kendall $\tau$, NED \\
proScript \citep{sakaguchi2021proscript} & crowd & 6{,}400 & \ding{51} & \ding{55} & \ding{55} & $\circ$ & edge F1 \\
How2Bench \citep{how2everything2026} & open web & 7{,}000 & \ding{51} & \ding{55} & \ding{55} & \ding{55} & LLM judge (failures) \\
\midrule
\headrow \bench{} (ours) & wikiHow, expert-prov. & 6,885 & \ding{51} & \ding{51} & \ding{51} & \ding{51} & accuracy, NED, judge \\
\bottomrule
\end{tabular}
\caption{Selected benchmarks for task decomposition or procedural
comprehension. \emph{Gen.}: the model must
produce step text. \emph{Miss.}: the model must identify a step withheld
from a given procedure. \emph{Err.}: the model must localise an incorrect
step injected into an otherwise correct procedure. \emph{Ord.}: the model
must recover or judge step order --- $\circ$ marks benchmarks that predict
pairwise dependencies over gold steps rather than requiring a full ordering.
Among the task-decomposition benchmarks in this table, none evaluates
recovery of missing or detection of incorrect steps, and none evaluates
comprehension and generation over a shared procedure pool. Sizes are as
reported by each paper; \bench{} is the count of curated procedures, each of
which yields an instance of every task.}
\label{tab:related}
\end{table*}

\subsection{Procedural Understanding}
Task-decomposition benchmarks generally measure whether a model can generate a suitable collection of steps for a goal. \citet{zhang2021decompose} use reference-overlap metrics for generated decompositions; TaskLAMA \citep{yuan2023tasklama} evaluates generated task graphs and separately probes pairwise temporal dependencies; TaskBench \citep{shen2024taskbench} embeds decomposition in a tool-automation pipeline; and UltraTool \citep{huang2024ultratool} evaluates natural-language plans before tools are used. PLANET \citep{li2025planet} collects multiple planning evaluations under one framework, further illustrating the breadth of behaviours subsumed by a single ``planning'' label. These benchmarks motivate evaluating planning before execution, but their plan-production scores do not localise a model's failure mode. Benchmarks that do localise it restrict the setting to where an oracle can check each sub-skill. Embodied Agent Interface \citep{DBLP:conf/nips/LiZWWZSGLLZLL0M24} factors embodied decision making into goal interpretation, subgoal decomposition, action sequencing, and transition modelling, using simulators to type errors as missing, additional, or misordered steps; ACPBench \citep{DBLP:conf/aaai/Kokel0SS25} poses atomic reasoning tasks over thirteen formal planning domains, with generative variants in ACPBench Hard \citep{DBLP:conf/iclr/KokelKSS26}; and WorfBench \citep{DBLP:conf/iclr/QiaoFQW0JXHC25} scores workflows by subsequence and subgraph matching, finding models markedly weaker on graph-structured plans than linear ones. Each depends on a simulator, a formal domain, or an annotated graph. \bench{} obtains a comparable per-capability readout from controlled perturbations of verified references instead.

Several works isolate parts of procedural understanding. CaT-Bench \citep{lal2024cat} evaluates whether a step should precede another in recipe plans, and \citet{anika2025ordering} reconstruct shuffled recipe sequences. Most closely, \citet{zhang2020reasoning} derive wikiHow tasks for goal-step inference and pairwise temporal ordering, using semantic similarity to obtain challenging negative examples. \bench{} inherits this controlled-inference tradition but evaluates gaps and complete orders within procedures, adds plausible inserted faults, and connects all of these diagnostics to generation over the same goals. PARADISE \citep{uzunoglu2024paradise} evaluates warnings and tips associated with procedures, \citet{yuan2023distilling} distil script knowledge for constrained generation, and proScript \citep{sakaguchi2021proscript} models partially ordered scripts. These tasks assess relations among supplied steps, but do not present a complete procedure with a topically plausible violation and ask the model to identify it. Error localisation of that kind exists where the answer is formally checkable: ProcessBench \citep{DBLP:conf/acl/ZhengZZLLYLZL25} asks for the earliest erroneous step in expert-annotated mathematical solutions, and DeltaBench \citep{DBLP:conf/acl/HeLLWB0P0ZS025} for faulty sections in long reasoning traces. The capability matters because models audit their own output unreliably: self-critique degrades plans in closed domains rather than repairing them \citep{DBLP:journals/corr/abs-2310-08118,DBLP:conf/iclr/StechlyVK25}, which argues for measuring detection as a capability of its own rather than assuming generation implies it. \bench{} supplies that diagnostic setting through incorrect-step detection and full-order recovery, while its missing-step tasks assess recovery with the surrounding procedural context visible.

Executable planning benchmarks provide another useful point of comparison. \citet{valmeekam2022large} and PlanBench \citep{valmeekam2023planbench} use formal action models that can determine whether a plan is valid. Their closed action spaces provide an execution oracle, whereas \bench{} evaluates open-domain procedures for which executing every candidate plan would require diverse real-world environments. The two settings therefore answer different questions: executable benchmarks assess success under a formal dynamics; \bench{} diagnoses how a model represents the content and order of a procedure without executing it.

\subsection{References and Generation Scoring}
Procedural corpora make different trade-offs among scale, domain, and source characteristics. wikiHow has supported summarisation, goal reasoning, and procedure discovery \citep{koupaee2018wikihow,zhang2020reasoning,lyu2021goal,zhou2022show}; recipes and instructional videos provide narrower domains \citep{yamakata2020english,tang2019coin}; proScript uses crowd-authored scripts \citep{sakaguchi2021proscript}; and How2Everything \citep{how2everything2026} mines web-scale procedures. Because \bench{} creates controlled faults from its references, it records editorial signals for every source article and validates its model-assisted curation, rather than treating source provenance as a guarantee of correctness.

For open-ended plan generation, ROUGE and BERTScore remain common points of comparison \citep{zhang2021decompose,shen2024taskbench,brahman2024plasma}. Judge-based protocols instead assess properties of a plan that need not match a single reference closely \citep{zheng2023judging}. The judges themselves are now measured: JudgeBench \citep{DBLP:conf/iclr/TanZMTC0PS25} finds strong judges near chance on response pairs with objectively correct answers, \citet{DBLP:journals/corr/abs-2406-12624} show high raw agreement coexisting with weak chance-corrected alignment, and \citet{DBLP:conf/iclr/DornerNH25} bound how much ground-truth labelling a judge can save once the judged models rival it. UltraTool evaluates six dimensions of a plan with an LLM judge \citep{huang2024ultratool}; How2Everything asks whether a plan contains a critical failure that prevents the goal being achieved \citep{how2everything2026}. \eval{} adopts the latter formulation for \bench{}'s generation tasks.
\section{Prompts}
\label{app:prompts}

All prompts are reproduced verbatim. Placeholders in braces are filled at runtime with the corresponding instance fields.

\subsection{Data Construction Prompts}
\label{app:pipeline-prompts}

These two prompts implement the pipeline stages in \S\ref{sec:data-pipeline}. All use GPT-4.1.

\begin{promptbox}{Stage 1: Task Filtering}
[System]
You are an expert in wikiHow task understanding and evaluation.

[User]
I have a method and its steps from wikiHow. Please analyse the method title and its steps and strictly evaluate whether they constitute a well-defined task. A task is a specific objective that can be completed by following a structured sequence of steps. Steps are the individual actions that, when executed in order, should lead to the successful completion of the task. Specifically, determine if completing the steps logically leads to accomplishing the task described in the method title, and if they can be modelled as a task and steps in a general.

Here is the method and its steps: {method}.

Please strictly follow the instruction, and output 'Yes' if it is logically a task and 'No' if it is not. Please do not provide any extra information or assumptions.
\end{promptbox}
\begin{promptbox}{Stage 2: Step Curation}
[System]
You are an expert in analysing procedural tasks.

[User]
Given a task title, the list of steps, and the description of each step, evaluate them based on the following criteria:
1. Necessity Check: Determine if each step is STRICTLY required to accomplish the task. If any step is unnecessary, REMOVE it.
2. Order Check: If the remaining steps are not in the correct sequential order, reorder them logically.

Here is the task title, the list of steps, and the description of each step: {method}.

Please follow the following instructions carefully:
- Remove the unnecessary steps from the initial steps.
- If reordering is needed for the remaining STRICTLY required steps, return the step titles arranged logically ordered.
- Respond strictly in JSON format as follows: {'Step 1': 'Title', 'Step 2': 'Title', ...}
- Please do not provide any details, just provide the final STRICTLY logically ordered steps in the provided JSON format.
\end{promptbox}

\subsection{Task Prompts}
\label{app:task-prompts}

These are the prompts given to evaluated models, one per task (\S\ref{sec:benchmark}).

\tcbinputlisting{
  sbprompt,
  title={T1: Last-Step Selection},
  listing file={prompts/task1-last-step-selection.txt}
}

\tcbinputlisting{
  sbprompt,
  title={T2: Missing-Step Recovery},
  listing file={prompts/task2-missing-step-recovery.txt}
}

\tcbinputlisting{
  sbprompt,
  title={T3: Incorrect-Step Detection},
  listing file={prompts/task3-incorrect-step-detection.txt}
}

\tcbinputlisting{
  sbprompt,
  title={T4: Step Reordering},
  listing file={prompts/task4-step-reordering.txt}
}

\tcbinputlisting{
  sbprompt,
  title={T5: Next-Step Generation},
  listing file={prompts/task5-next-step-generation.txt}
}

\tcbinputlisting{
  sbprompt,
  title={Full Plan Generation},
  listing file={prompts/task6-full-plan-generation.txt}
}

\paragraph{Decoding.}
All runs use greedy decoding ($T=0$) unless a model's provider requires otherwise. We add one sampled configuration, Gemini~3 Flash at $T=1$, to check that our conclusions are not artefacts of greedy decoding. Models supporting extended reasoning are run in both modes where available (Qwen3~32B); Command~A~Reasoning and Gemini~3 Flash are evaluated in reasoning mode only, as their non-reasoning counterparts are separate models rather than a toggle.

\paragraph{Parsing.}
Model output is parsed by a per-task extractor that attempts strict JSON first, then a progressively more tolerant sequence: fenced-block stripping, JSON5, Python literal evaluation, and finally a regular-expression fallback for the expected fields. Outputs that survive none of these are recorded as unparsed and scored as failures.
\section{Corpus Details}
\label{app:data}

\subsection{Datasheet}
\label{app:datasheet}

\paragraph{Motivation.}
\bench{} was created to evaluate whether language-model agents can understand, verify, and construct real-world task procedures. No existing resource measured comprehension and generation on a shared pool of procedures with a stated reference provenance.

\paragraph{Composition.}
\numsamples{} procedures, each a goal plus an ordered list of reference steps, derived from \numbencharticles{} wikiHow articles across \numcats{} categories. \numsteps{} reference steps in total. Instances carry a category label and a stable identifier linking them across all \numtasks{} tasks. Text is English.

\paragraph{Collection.}
Articles were crawled from wikiHow in February 2025. The crawl retained article text, method structure, and the article's editorial metadata (expert-authorship flag, editorial review tag, co-author count, reference count, view count, reader helpfulness rating). Processing is described in \S\ref{sec:data-pipeline}.

\paragraph{Preprocessing.}
Two GPT-4.1 stages: task filtering and step curation (prompts in Appendix~\ref{app:pipeline-prompts}). Raw crawl, intermediate outputs, and final pool are all released, so any stage can be inspected or re-run.

\subsection{Provenance Signals in Detail}
\label{app:provenance}
wikiHow attaches two independent quality signals to articles. The
\emph{expert co-author} flag indicates the article was written with a named, credentialed subject expert. The \emph{editorial tag} records a separate review process with three levels: \emph{Expert Approved} (a subject expert reviewed the finished article), \emph{Fact Checked} (staff verified factual claims against cited sources), and \emph{Tested} (staff physically performed the procedure). Table~\ref{tab:corpus} reports the distribution over \bench{}'s source articles. Neither signal is a correctness proof, and we do not use them to filter: every article that passes the pipeline enters the benchmark regardless of its tags. They are reported because they characterise the population the references are drawn from, and because that population differs materially from an unfiltered web crawl.

\begin{figure*}[t]
    \centering
    \includegraphics[width=\textwidth]{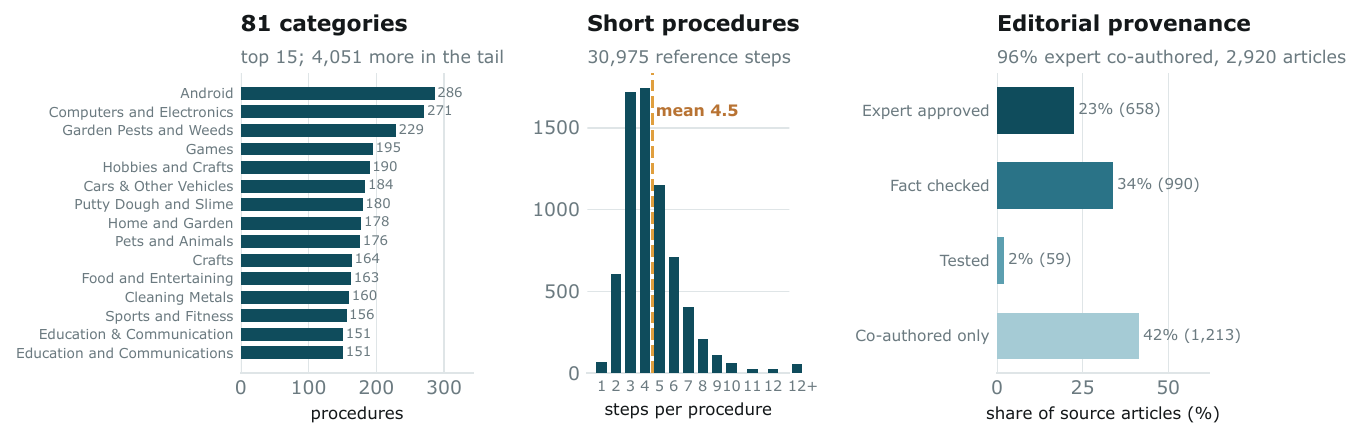} \caption{\bench{} corpus. \emph{Left}: procedures per category, top 15 of \numcats{}; the distribution is long-tailed because how-to content is. \emph{Centre}: procedures are short --- a median of 4 steps --- which is what makes exact-match reordering a demanding but not hopeless task. \emph{Right}: editorial provenance of the source articles.}
    \label{fig:corpus}
\end{figure*}

\subsection{Human Validation}
\label{app:data-human-eval}
Stages 1 and 2 make the consequential decisions in our pipeline: Stage 1 determines which candidates qualify as procedural tasks, while Stage 2 determines the reference steps used to construct the benchmark. We therefore evaluate both stages against independent human judgements.

For each stage, we sample \numannotated{} instances and collect independent judgements from \numannotators{} annotators. Annotators receive the same instruction as the model and answer the corresponding evaluation question without seeing its decision. We construct the two samples differently to reflect what each validation is intended to measure. For Stage 1, we sample methods by including candidates that the Stage 1 filter accepts and rejects to evaluate whether the filtering criterion generalises beyond the procedures retained in the benchmark. For Stage 2, we sample directly from \bench{}, since the objective is to assess the quality of the reference procedures used in the final benchmark.

\paragraph{Task Filtering.}
Annotators judge whether each candidate constitutes a well-defined procedural task. Human annotators agree with one another on 84.5-87.0\% of instances, while their majority judgment agrees with the model on 82.0\%.

\paragraph{Step Curation.}
Annotators are shown the original and curated step sequences side by side and judge whether the curation preserves all necessary steps, removes unnecessary ones, and produces a logical ordering. Human annotators agree with one another on 84.5--89.5\% of instances and judge 81.5\% of the resulting curations to be correct.

\begin{table*}[t]
\centering
\small
\begin{tabular}{l r l r}
\toprule
\headrow \multicolumn{2}{c}{\textbf{Scale}} & \multicolumn{2}{c}{\textbf{Provenance of source articles}} \\
\cmidrule(lr){1-2}\cmidrule(lr){3-4}
Procedures & 6,885 & Distinct articles & 2,920 \\
Reference steps & 30,975 & Expert co-authored & 2,806 (96.1\%) \\
Categories & 81 & Editorially reviewed & 1,707 (58.5\%) \\
Steps per procedure & 4.5 & \quad Fact checked & 990 \\
\quad median (range) & 4 (1--28) & \quad Expert approved & 658 \\
Words per step & 7.27 & \quad Tested & 59 \\
Words per task title & 10.56 & Median readership & 244,916 views \\
Largest / smallest category & 286 / 2 & Cumulative readership & 1.49\,B views \\
Median category size & 68 & Median helpfulness & 82\% \\
\bottomrule
\end{tabular}
\caption{\bench{} at a glance. \emph{Left}: benchmark scale.
\emph{Right}: provenance of the wikiHow articles the procedures are drawn
from. wikiHow attaches an authorship record to every article; the great
majority of \bench{}'s sources were co-authored with a credentialed subject
expert, and a further editorial process fact-checks, expert-approves, or
physically tests a large subset. Readership and helpfulness ratings are
reader-supplied signals that these are procedures people actually follow.}
\label{tab:corpus}
\end{table*}

\section{\eval{} Details}
\label{app:stepeval}

\subsection{Judge Prompt and Codebook}
\label{app:judge-prompt}

The judge receives the goal, the reference procedure $S^\star$ (labelled L1), and the candidate plan $\hat{S}$ (labelled L2), and returns JSON with a free-text \texttt{reasoning} field and a list of \texttt{critical\_failures}, each linked to the reference and candidate steps it concerns.

\paragraph{Failure taxonomy.}
\begin{itemize}[leftmargin=1.4em,itemsep=1pt,topsep=3pt]
  \item \textbf{Contradiction of the goal} --- a step directly contradicts a condition stated in the goal.
  \item \textbf{Contradiction of the reference} --- a step diverges from a reference step in a way that prevents the goal being achieved.
  \item \textbf{Internal inconsistency} --- a step is inconsistent with another step in the same plan.
  \item \textbf{Incoherence} --- the plan has very low readability or logical flow; determinable without consulting the reference.
  \item \textbf{Severe vagueness} --- the plan omits so much essential detail that it is unusable as instructions.
  \item \textbf{Missing critical action} --- an essential reference step is absent with no equivalent or implied action.
  \item \textbf{Extraneous action} --- an added step is unnecessary, confusing, or counterproductive.
  \item \textbf{Redundant repetition} --- a step repeats an earlier one where the reference has no such repetition.
\end{itemize}

\paragraph{Acceptable variation.}
The codebook explicitly instructs the judge \emph{not} to flag differences of tone, phrasing, or level of detail; differences of emphasis or ordering that do not change the outcome; additional steps that are neutral or practical; or omissions that are reasonably implied by another step. This is what distinguishes the protocol from generic preference judging, and it is why the protocol can credit a plan that is correct by a different route.

\paragraph{Aggregation.}
A plan is scored \textsf{no\_failure} if the returned failure list is empty, \textsf{has\_failure} otherwise. The valid-plan rate is the share scored \textsf{no\_failure}. Following \citet{how2everything2026}, we take ``any failure'' rather than ``first failure'' as the aggregation, which yields higher inter-annotator agreement in their validation.

\subsection{Judge Configurations}
\label{app:judge-config}

\begin{itemize}[leftmargin=1.4em,itemsep=1pt,topsep=3pt]
  \item \textbf{GPT-5} \citep{openai2026gpt5}: Reference judge used to evaluate selected models and provide annotations for distillation; it is not used for the final reported evaluation. Default temperature, with no \texttt{top\_p} or \texttt{top\_k} settings, in non-thinking mode.
  \item \textbf{\eval{}}: Qwen3-8B fine-tuned for three epochs on GPT-5 annotations. This judge provides all final reported evaluations for both \Tnext{} and full-plan generation across all configurations and both datasets. Decoding settings: $T=0$, \texttt{top\_p}~0.95, and \texttt{top\_k}~20.
\end{itemize}

\subsection{Distillation Setup}
\label{app:distill}

Training data joins GPT-5 judge outputs back to their source instances using the same prompt-rendering code the judge harness uses at inference, so training and inference formats cannot drift. We fine-tune Qwen3-8B for 3 epochs, effective batch size 16, learning rate $1\times10^{-5}$, cosine schedule with 3\% warmup, bf16, gradient checkpointing, no sequence-length truncation.

\tcbinputlisting{
  sbprompt,
  title={StepEval Critical Failure},
  listing file={prompts/stepeval-prompt.txt}
}

\subsection{Applying \eval{} to \Tnext{}}
\Tnext{} asks for one step, not a procedure, so a plan-level judge does not apply to it directly. Rather than introduce a second judge with its own calibration, we reframe each \Tnext{} instance as a single-step splice: the reference procedure with only the step at the target index replaced by the model's generated step, every other position identical. Because reference and candidate differ at exactly one position, any failure the judge reports must trace to that step --- including failures that depend on context, such as redundancy with an earlier step or inconsistency with a later one, which a bare step-against-step comparison could not detect. It also avoids the spurious ``missing action'' flags that comparing a one-step output against a full reference would produce on every instance.

\begin{figure*}[t]
    \centering
    \includegraphics[width=\textwidth]{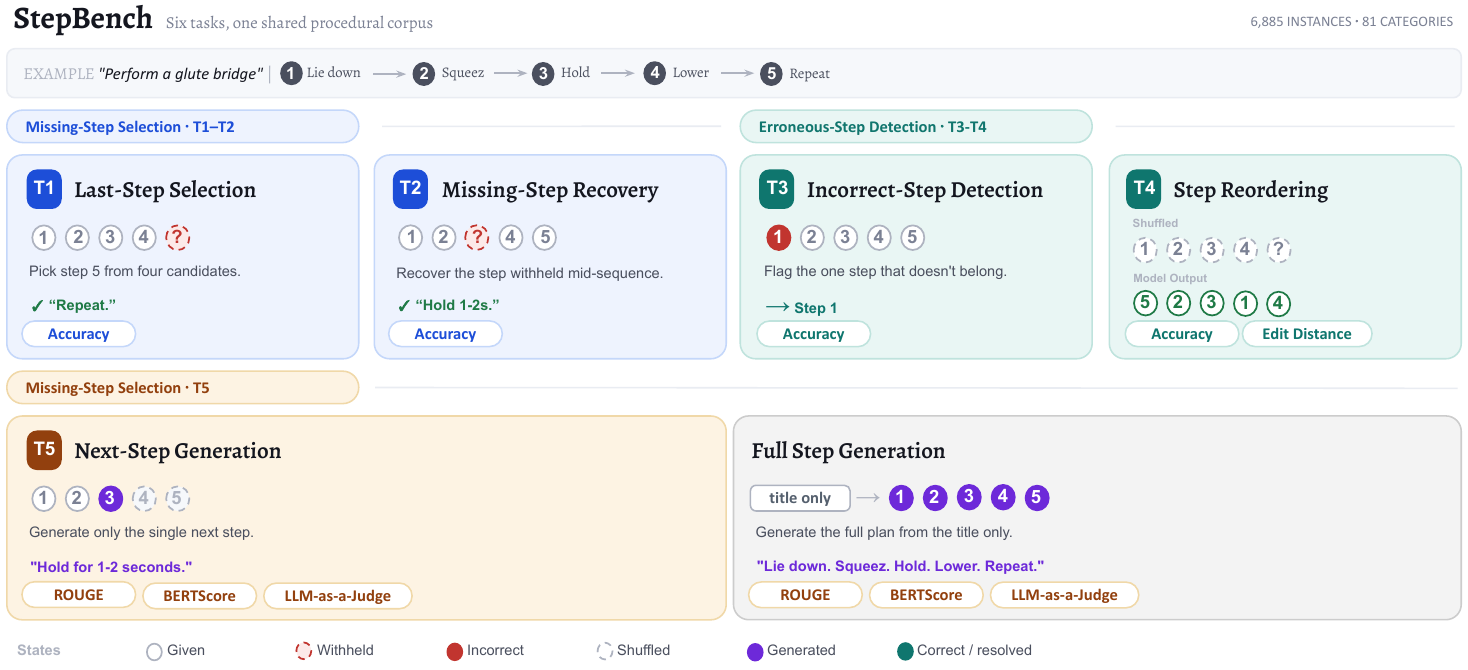}
    \caption{The five \bench{} diagnostic tasks and full-plan generation on a shared procedure (``perform a glute bridge''). \Tlast{}--\Tgap{} withhold a step for recovery from candidates; \Terr{}--\Tord{} introduce incorrect or shuffled steps for detection or reordering; and \Tnext{} requires generating the next step. Full-plan generation is shown separately as an end-to-end evaluation setting. Circles indicate each step's state: given, withheld, incorrect, shuffled, or generated.}
    \label{fig:stepbench}
\end{figure*}

\subsection{Validation via Human Annotations}
\label{app:stepeval-human-eval}
To validate the definition of critical failures for \eval{}, we ask three human annotators to label 200 examples using our evaluation protocol. We reuse the procedures sampled for step-curation evaluation, pairing each with a model-generated procedure. Annotators inspect the goal, reference procedure, and generated procedure and identify any critical failures. We derive a binary label, `has\_failure` if at least one critical failure is identified, and `no\_failure` otherwise.

\paragraph{Inter-annotator agreement.} With binary score aggregation, we observe Krippendorff's $\alpha = 0.612$. Given the non-executable, open-world setting, we do not expect near-perfect agreement. To contextualise judge performance, we also measure leave-one-out agreement among annotators, which ranges from 87.2\% to 90.5\%, excluding examples where the other two annotators disagree.

\paragraph{Evaluating GPT-5 against human labels.} We compare the judgements from GPT-5 with the human majority labels on the same 200 examples. GPT-5 achieves 85.5\% agreement overall, with 87.1\% agreement on human-majority \textit{has\_failure} cases and 83.8\% on \textit{no\_failure} cases. We interpret these results alongside human leave-one-out agreement to assess how closely GPT-5 reproduces human judgements under the critical-failure protocol.

\section{Additional Diagnostic Results}
\label{app:results}

This appendix provides further breakdowns for the five diagnostic tasks and their relationship with full-plan generation. We use the same scores as Section~\ref{sec:results}: accuracy for \Tlast{}--\Tord{}, normalised edit distance for partial ordering agreement, and \eval{} VPR for \Tnext{} and full-plan generation.

\subsection{Performance by Procedure Length}
\label{app:length}

\begin{figure*}[ht]
    \centering
    \includegraphics[width=\textwidth]{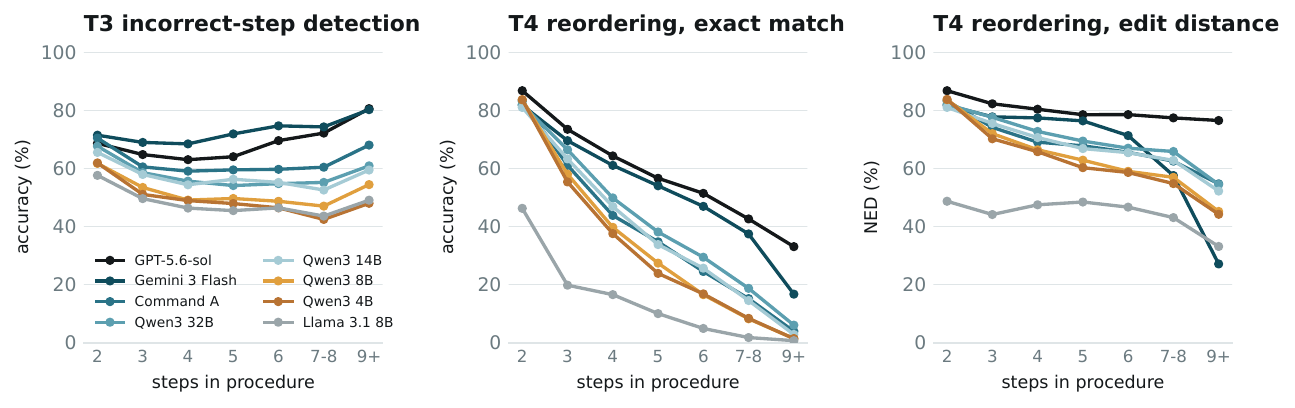}
    \caption{Scores by procedure length for eight configurations. Left: incorrect-step detection accuracy. Centre: exact-order accuracy. Right: normalised edit distance, where higher scores indicate closer reference agreement. Gemini~3 Flash uses $T=0$; Qwen3~32B uses non-thinking mode.}
    \label{fig:length}
\end{figure*}

Figure~\ref{fig:length} shows incorrect-step detection and reordering scores by procedure length for eight configurations. Detection does not follow the same pattern across models. Gemini~3 Flash's accuracy rises from 68.5\% at four steps to 80.3\% at nine or more, while Qwen3~4B falls from 51.1\% at three steps to 42.5\% at seven or eight. These comparisons show different responses to longer procedures; they do not establish whether additional context itself causes the change, since the procedures also differ.

Exact-order accuracy declines with length. GPT-5.6-sol achieves 86.8\% at two steps and \bestTfournine{}\% at nine or more. Its normalised edit distance score changes less, from 86.8 to 76.6. Reporting both metrics therefore distinguishes complete reference-order recovery from partial agreement. Neither score identifies the particular dependency that was violated, and a different order can still be valid when actions can be exchanged. The length breakdown describes difficulty in recovering the reference sequence rather than directly measuring execution success.

\subsection{Position of the Missing or Incorrect Step}
\label{app:position}

\begin{figure*}[t]
    \centering
    \includegraphics[width=\textwidth]{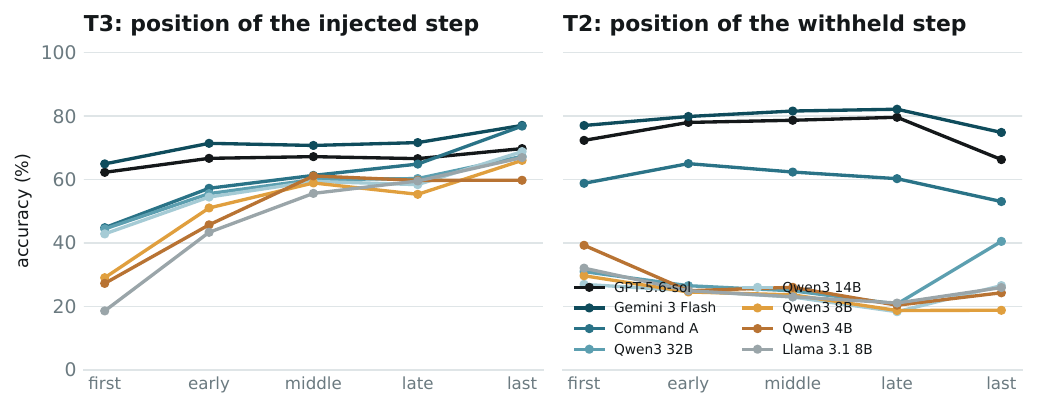}
    \caption{Accuracy by relative step position for the same eight configurations as Figure~\ref{fig:length}. Left: the replaced step in \Terr{}. Right: the missing step in \Tgap{}. Position effects vary across tasks and models.}
    \label{fig:position}
\end{figure*}

Figure~\ref{fig:position} shows accuracy by the relative position of the missing or incorrect step. Neither sequence boundary offers a general advantage. On \Terr{}, Gemini~3 Flash scores 64.9\% when the incorrect step is first and 77.0\% when it is last. Qwen3~4B scores 27.2\% and 59.7\%, respectively. On \Tgap{}, the pattern differs: Gemini~3 Flash scores 81.5\% for a middle gap, compared with 77.0\% for the first position and 74.8\% for the last. Qwen3~4B instead performs best at the first position (39.2\%).

These differences show why a single task score does not fully describe a model's recovery or detection performance. Position groups also contain different procedures and length distributions, so the results should not be read as evidence of a particular attention mechanism.

\subsection{Per-Category Results}
\label{app:percategory}

\begin{figure*}[t]
    \centering
    \includegraphics[width=0.8\textwidth]{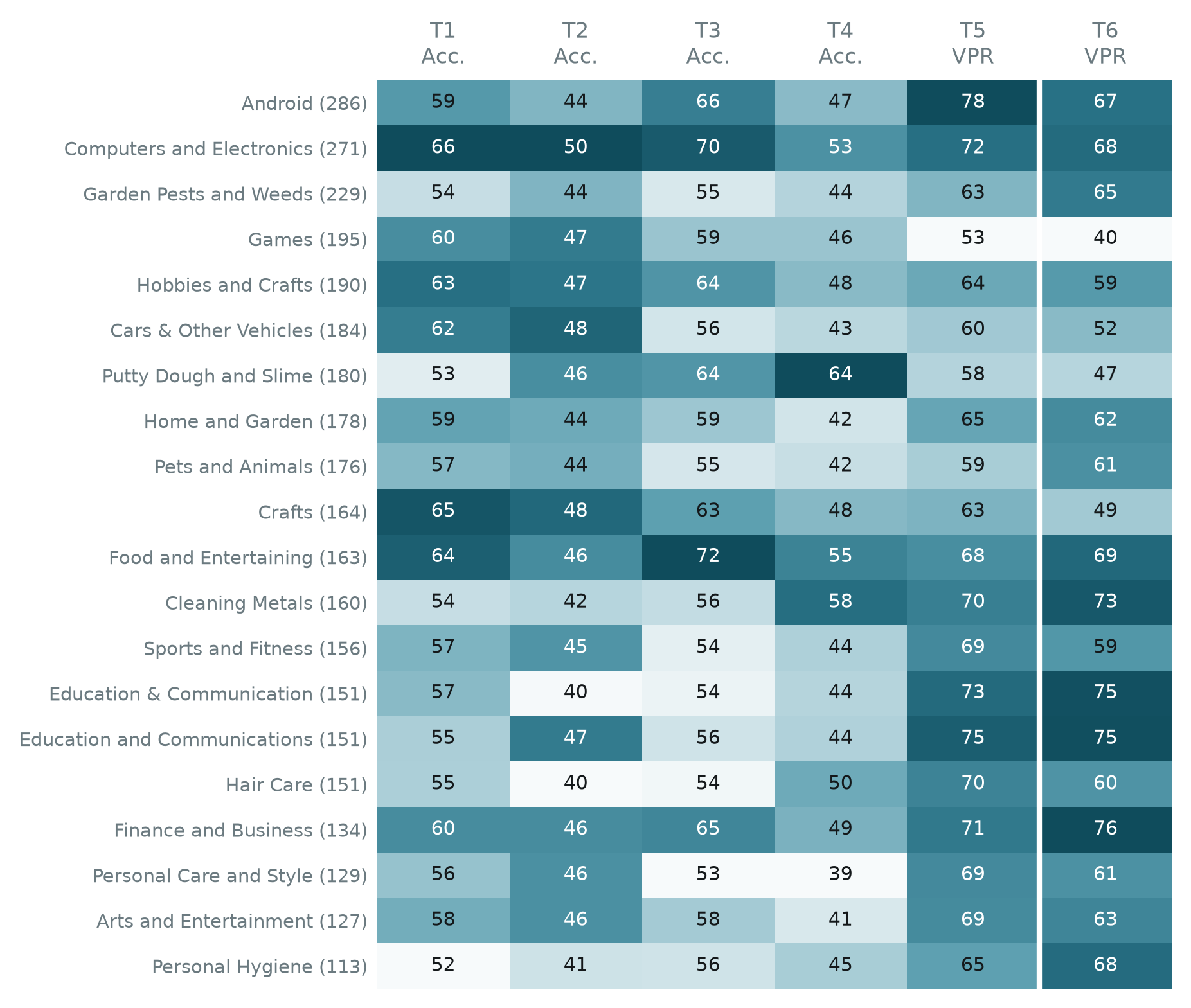}
    \caption{Mean scores across \numruns{} configurations for the twenty largest categories, ordered by procedure count. Counts in parentheses refer to the source procedure pool; task-specific counts can differ. \Tlast{}--\Tord{} use accuracy and \Tnext{} and full-plan generation use VPR. Shading is normalised within each column.}
    \label{fig:category}
\end{figure*}

Figure~\ref{fig:category} shows mean scores across all \numruns{} configurations for the twenty largest categories. It includes all five diagnostic tasks alongside full-plan generation. Categories are ordered by procedure count, and shading compares categories within each task. We do not combine accuracy and VPR into a single category-difficulty score.

These breakdowns help identify domains in which a diagnostic task is particularly difficult. However, differences may reflect procedure length, content, or the available models' familiarity with a domain. They do not by themselves establish that one domain has stricter dependencies or more interchangeable actions than another. The figure shows the largest categories for readability; results for all \numcats{} categories are available in the released per-category analysis.

\subsection{Sensitivity of Diagnostic-to-Generation Correlations}
\label{app:diagnostic-correlations}

Table~\ref{tab:diagnostic-sensitivity} shows how the correlations in Section~\ref{sec:results-proxy} change under two exclusions. First, we keep one configuration per underlying model by retaining Gemini~3 Flash at $T=0$ and Qwen3~32B without thinking. Second, we exclude GPT-5.6-sol from the full set to check whether the strongest full-plan result drives the association. All five correlations remain positive and above 0.80 in both checks. This supports the overall association, although the ordering of the diagnostic tasks by correlation changes slightly.

These are sensitivity checks on the same model sample, not independent validation. Models within a family still share architecture and training, and the results do not establish prediction accuracy for individual plans or for models outside this sample.

\begin{table}[t]
\centering
\small
\adjustbox{max width=\columnwidth}{%
\begin{tabular}{l r r r r r r}
\toprule
\headrow \textbf{Configurations} & $n$ & \Tlast{} & \Tgap{} & \Terr{} & \Tord{} & \Tnext{} \\
\midrule
All configurations & 14 & 0.95 & 0.85 & 0.93 & 0.96 & 0.89 \\
One per model & 12 & 0.96 & 0.82 & 0.93 & 0.94 & 0.86 \\
Without GPT-5.6-sol & 13 & 0.96 & 0.82 & 0.93 & 0.95 & 0.87 \\
\bottomrule
\end{tabular}}
\caption{Spearman correlations between diagnostic scores and full-plan VPR. Each column uses accuracy for \Tlast{}--\Tord{} or VPR for \Tnext{}. The two restricted sets are sensitivity checks, not independent test sets.}
\label{tab:diagnostic-sensitivity}
\end{table}

\subsection{Sampling Temperature}
\label{app:temperature}

Table~\ref{tab:main} shows results for Gemini~3 Flash at $T=0$ and $T=1$. The two configurations differ by at most 1.5 percentage points across \Tlast{}--\Tord{} accuracy and VPR for \Tnext{} and full-plan generation. Full-plan VPR is 84.5\% and 83.0\%, respectively. Partial-order agreement changes more: normalised edit distance is 73.4 at $T=0$ and 76.6 at $T=1$. The $T=1$ configuration also has fewer unparseable reordering outputs (57 versus 440), while exact-order accuracy changes from 58.7\% to 59.4\%. These results show that the two settings produce similar headline scores, but differ in output formatting and partial ordering agreement.

\subsection{Results and Controls on \bench{}-Val}
\label{app:contam}

Table~\ref{tab:contam-runs} shows the change for each configuration after restricting the main set to the \numsharedcats{} shared categories and reweighting its scores to the validation set's procedure-length distribution. The length groups are two--three, four--five, six--eight, and nine or more steps. These adjustments reduce differences in category coverage and length; they do not make the two sets equivalent in every respect.

Twelve of \numruns{} configurations change by less than three points on \Tlast{}. The largest decrease is Qwen3~32B without thinking (\contamqwenTone{} points), compared with \contamqwenthinkTone{} for its thinking configuration. In contrast, every configuration has lower scores on \Tord{}, \Tnext{}, and full-plan generation. Thus, the small average changes on selection, recovery, and detection should not be generalised to all five diagnostic tasks.

\begin{table*}[t]
\centering\small
\begin{tabular}{l r r r r r r}
\toprule
\headrow & \multicolumn{5}{c}{\textbf{Diagnostic tasks}} & \textbf{Full plan} \\
\cmidrule(lr){2-6}\cmidrule(lr){7-7}
\headrow \textbf{Configuration} & \Tlast{} & \Tgap{} & \Terr{} & \Tord{} & \Tnext{} (\vpr{}) & \vpr{} \\
\midrule
GPT-5.6-sol (T=0, no think) & -0.7 & \weak{-4.0} & +0.6 & \weak{-11.7} & \weak{-8.3} & \weak{-15.7} \\
Gemini 3 Flash (think, T=0) & -0.2 & -0.6 & +1.9 & \weak{-20.7} & \weak{-8.0} & \weak{-13.8} \\
Gemini 3 Flash (think, T=1) & -2.9 & -1.2 & -0.3 & \weak{-20.1} & \weak{-10.7} & \weak{-14.2} \\
Command A Reasoning (think) & -1.7 & -2.3 & \weak{-3.3} & \weak{-10.4} & \weak{-12.1} & \weak{-16.0} \\
Command A & -1.3 & +1.6 & +0.5 & \weak{-7.5} & \weak{-7.7} & \weak{-13.9} \\
Qwen3 32B (think) & -2.3 & +4.8 & +0.4 & \weak{-11.1} & \weak{-9.9} & \weak{-22.0} \\
Qwen3 32B & \weak{-8.8} & +1.7 & +0.1 & \weak{-7.2} & \weak{-7.9} & \weak{-21.7} \\
Gemma 3 27B & +0.0 & -2.2 & -1.1 & \weak{-5.8} & \weak{-7.8} & \weak{-13.0} \\
Qwen3 14B & +0.7 & +1.2 & -0.7 & \weak{-3.8} & \weak{-6.7} & \weak{-22.2} \\
GLM-4 9B & \weak{-3.6} & +0.6 & \weak{-5.0} & \weak{-5.5} & \weak{-3.5} & \weak{-13.7} \\
Qwen3 8B & -1.1 & -1.6 & +2.3 & \weak{-5.8} & \weak{-4.0} & \weak{-19.0} \\
Llama 3.1 8B & +0.4 & -1.6 & \weak{-5.5} & -0.1 & \weak{-6.1} & \weak{-22.1} \\
Qwen3 4B Instruct (2507) & +0.1 & -1.7 & +3.6 & \weak{-4.8} & \weak{-6.6} & \weak{-17.3} \\
Qwen3 4B & +1.9 & -2.0 & -2.0 & -2.2 & \weak{-5.7} & -2.0 \\
\midrule
\textbf{Mean} & \textbf{-1.4} & \textbf{-0.5} & \textbf{-0.6} & \textbf{-8.3} & \textbf{-7.5} & \textbf{-16.2} \\
\bottomrule
\end{tabular}
\caption{Per-configuration change from \bench{} to the 2026
validation set, after matching category composition and reweighting to the
2026 procedure-length distribution. \Tlast{}--\Tord{} use accuracy;
\Tnext{} and full-plan generation use StepEval VPR. Values are percentage-point changes.
Every configuration has lower scores on reordering, next-step generation,
and full-plan generation. Amber marks a decrease of more than three points.}
\label{tab:contam-runs}
\end{table*}

\paragraph{Remaining differences between the datasets.}
The validation set contains only \numtwosix{} procedures and uses a variant of the construction pipeline described in Section~\ref{sec:data-2026}. Its goal statements are also shorter on average (\goalwordstwosix{} words versus \goalwordsmain{}). A separate sensitivity check matches shared categories and goal length instead of procedure length. The mean full-plan VPR change is then \goaldeltaTplan{} points, compared with \deltaTplan{} under the main adjustment. This check is limited by sparse support: the main set has only \mainshortgoals{} goals of six words or fewer, compared with \twosixshortgoals{} in the validation set.

\paragraph{Limits of the temporal comparison.}
A lower score on newer procedures is not sufficient evidence of memorisation. The model sample, dataset construction, and task difficulty can also affect the differences. GPT-5.6-sol's training chronology is unverified, and documented cutoffs for other models do not measure their exposure to individual articles. We therefore interpret these results as a check of how diagnostic profiles change across the two datasets, rather than as a test that rules contamination in or out.

\section{Experimental Setup}
\label{app:experiment}

Table~\ref{tab:model-cutoffs} lists the evaluated models, their exact identifiers, and the dates used for temporal comparisons. Our evaluation covers \numruns{} configurations of \nummodels{} models, including open-weight and proprietary models. The Qwen3 family spans four parameter sizes, supporting comparisons within a model family, while additional configurations examine reasoning mode and sampling temperature. Where providers disclose a knowledge cutoff, we report it directly; otherwise, we use the model's release date as a conservative temporal bound, rather than as its training cutoff. GPT-5.6-sol is included in the performance evaluation but excluded from temporal-exclusion claims because its relevant dates are unverified.

\begin{table*}[t]
\centering
\small
\adjustbox{max width=\textwidth}{%
\begin{tabular}{llll}
\toprule
\headrow \textbf{Model} & \textbf{Model ID} & \textbf{Knowledge Cut-off} & \textbf{Date basis} \\
\midrule
Qwen3-4B & \texttt{Qwen/Qwen3-4B} & 2025-04-29 & Release date \\
Qwen3-8B & \texttt{Qwen/Qwen3-8B} & 2025-04-29 & Release date \\
Qwen3-14B & \texttt{Qwen/Qwen3-14B} & 2025-04-29 & Release date \\
Qwen3-32B & \texttt{Qwen/Qwen3-32B} & 2025-04-29 & Release date \\
Qwen3-4B-Instruct-2507 & \texttt{Qwen/Qwen3-4B-Instruct-2507} & 2025-08-06 & Release date \\
Llama-3.1-8B-Instruct & \texttt{meta-llama/Llama-3.1-8B-Instruct} & 2023-12 & Knowledge cutoff \\
Gemma-3-27B-IT & \texttt{google/gemma-3-27b-it} & 2025-03-12 & Release date \\
GLM-4-9B & \texttt{zai-org/GLM-4-9B-0414} & 2025-04-15 & Release date \\
GPT-5.6-sol & \texttt{gpt-5.6-sol} & Unverified & Not used for temporal claims \\
Gemini 3 Flash & \texttt{gemini-3-flash-preview} & 2025-01 & Knowledge cutoff \\
Command A & \texttt{command-a-03-2025} & 2024-06-01 & Knowledge cutoff \\
Command A Reasoning & \texttt{command-a-reasoning-08-2025} & 2024-06-01 & Knowledge cutoff \\
\bottomrule
\end{tabular}}
\caption{Evaluated models, exact identifiers, and knowledge cutoffs. Where the knowledge cutoff is undisclosed and a release date is available, we use that date as a proxy, as indicated in the final column. GPT-5.6-sol is excluded from temporal-exclusion claims because neither date is verified. Month-only dates preserve the precision of the reported cutoff.}
\label{tab:model-cutoffs}
\end{table*}

\end{document}